\documentclass[11pt]{article}

\usepackage[preprint]{acl}
\usepackage{times}
\usepackage{latexsym}
\usepackage[T1]{fontenc}
\usepackage[utf8]{inputenc}
\usepackage{microtype}
\usepackage{booktabs}
\usepackage{tabularx}
\usepackage{multirow}
\usepackage{amsmath}
\usepackage{tikz}
\usetikzlibrary{arrows.meta,positioning}
\hypersetup{
  pdftitle={Evidence First, Arithmetic Second: A System Report and Failure Analysis for DocSem}
}
\hypersetup{pdfauthor={Divya Godara and Sachin Gupta}}

\newcolumntype{Y}{>{\raggedright\arraybackslash}X}
\newcommand{\docsem}{\textsc{DocSem}}
\newcommand{\system}{\textsc{EviCalc}}

\title{Evidence First, Arithmetic Second: A System Report and Failure Analysis for \docsem\thanks{Accepted as a shared-task system paper at DocInsights 2026, co-located with EMNLP 2026. This is the authors' preprint version.}}

\author{Divya Godara \\
  Independent Researcher \\
  San Jose, USA \\
  \texttt{godaradivya@gmail.com} \\\And
  Sachin Gupta \\
  Independent Researcher \\
  San Jose, USA \\
  \texttt{sachinkg12@gmail.com}}

\begin{document}
\maketitle

\begin{abstract}
\system{}, our system for the \docsem{} shared task, achieved 8.61\% joint
accuracy on 1,730 tasks in the official final test evaluation. It reads a PDF,
selects a passage, asks a language model to write an arithmetic expression, and
evaluates that expression in local code. Saved intermediate results support
inspection of failures. A separate public-validation run achieved 92.17\%
answer accuracy and 1.00 evidence F1. The configurations and metrics differ,
so these scores are not a controlled comparison. Our manual, post-hoc analysis
is descriptive: in one inspected case, optical character recognition (OCR) and
block grouping merged the relevant passage into another block, and the system
answered from unrelated text. An exploratory study of reading page images on
100 documents returned evidence identifiers for only 22 documents. These
descriptive findings motivate further evaluation; they do not establish the
causes of the overall score.
\end{abstract}

\section{Introduction}

Document question answering requires finding the relevant passage before
reasoning about it. In \docsem{}, built on GSM-SEM \citep{singh2026gsmsem}, each
task provides an image-only PDF and a paraphrased query. A system
must return both a numeric answer and the printed identifier of the supporting
passage. We call that passage the evidence block. A correct calculation from
the wrong block fails the joint answer-and-evidence requirement.

\system{} obtained 8.61\% joint accuracy in the official test evaluation. We
report the submitted system and descriptive failure analyses. Saved passages,
calculations, and answers support case-level inspection, not estimates of each
error type's contribution to the overall score.

\paragraph{Relation to earlier work.}
DocVQA introduced a benchmark for question answering over document images
\citep{mathew2021docvqa}. Program-aided language
models delegate calculation to an interpreter after a model generates a program
\citep{gao2023pal}. Our system uses a restricted arithmetic expression for this
step. The contribution here is an account of how this design behaved on
\docsem{}, where both the answer and the evidence identifier matter.

\begin{figure*}[t]
\centering
\includegraphics[width=\textwidth]{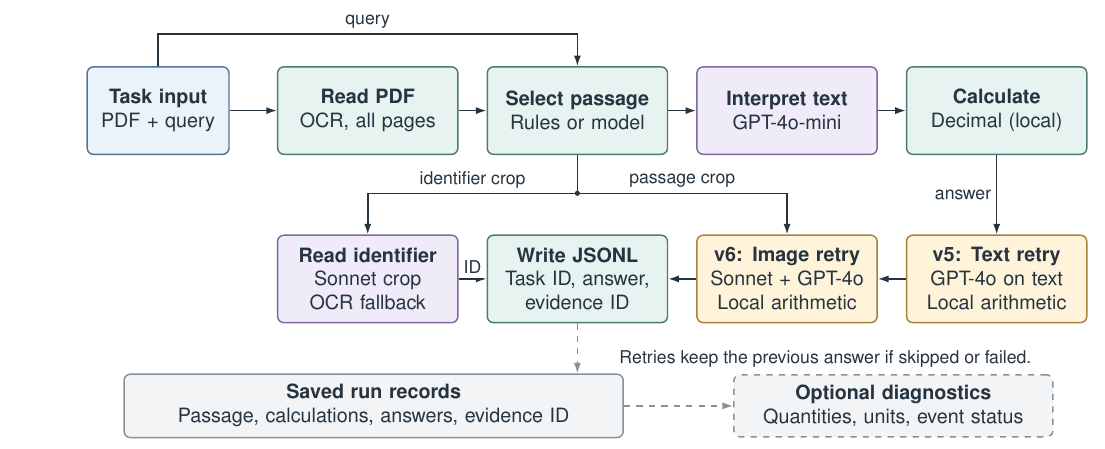}
\caption{Submitted adaptive v6 pipeline. Blue marks inputs, purple model calls,
teal selection/calculation and output, amber conditional retries, and gray
records and diagnostics. Arithmetic expressions are evaluated in local code.
Crops come from the selected passage in the supplied PDF.
The v5/v6 passes change only answers: unflagged rows and failed retries retain
the prior answer (Section~\ref{sec:submitted}). Dashed arrows lead to saved
records and optional checks, not a required verification step. Sonnet denotes
\texttt{claude-sonnet-5}.}
\label{fig:pipeline}
\end{figure*}

\section{Task and evaluation setting}

The participant release contains 908 labeled training tasks and 217 validation
tasks. Each task has an opaque instance identifier, a query, and a path to one
PDF. The test release contains 1,730 tasks. The instructions require systems to
solve from the supplied PDF rather than filenames, metadata, or an external source
lookup. The output is JSONL: an instance identifier, an answer string, and one or
more visible evidence block identifiers.

The public validation service reports answer accuracy and evidence scores. The
test service reports joint accuracy: the fraction of tasks for which both the
answer and the evidence are correct. Missing predictions count as wrong. The
validation answer score and test joint score are therefore different measures.

We used public training labels and validation feedback for development.
Predictions did not use organizer-held test answers, source solutions, or an
answer table indexed by task ID. Test analysis used released PDFs, extracted
text, saved runs, and aggregate submission feedback.

\section{System design}

Figure~\ref{fig:pipeline} shows the submitted test flow from task to prediction.
Each task names its PDF; the system does not search a collection of unrelated
documents. The reader, selector, model adapter, and calculator are replaceable
components. Their implementations and settings varied across experiments.

\subsection{Document reading and evidence selection}
\label{sec:reading}

The reader renders PDF pages as images. Tesseract OCR \citep{smith2007tesseract}
extracts text and its position on each page, and local code groups the lines into
blocks. Each block stores text, a page location, and an extracted identifier.
The public-validation selector scores blocks for problem-introduction phrases,
numbers, a question mark, and query words. It penalizes phrases associated with
administrative notes and returns the highest-scoring block.

On public data, the selector considered blocks \texttt{b06} to \texttt{b13} on
page 1. That assumption came from training documents and did not describe the
test layout. The submitted test version processed all pages and used a selector that
combined query words with topic words elsewhere in the document. Uncertain
selections could be referred to a language model choosing from a short list of
extracted blocks. It also cropped the selected block's opening and
used a vision model to reread its printed identifier. This changes the identifier
reading, not the choice of passage.

\subsection{Interpretation and arithmetic}

The model receives the query and selected evidence text. In the structured
configuration, it lists the requested quantity, facts quoted from the passage,
units, assumptions, and an arithmetic expression. The expression uses numeric
constants and a restricted set of operators and functions. Local Python code
checks the syntax and evaluates it with \texttt{Decimal}, using 40-digit working
precision. It rejects unsupported operations and non-finite results. It removes
insignificant trailing zeros when formatting the answer.

Computation is repeatable for a given expression, but the model can choose the
wrong quantities or operations. Optional rules check quantities, units, repeated
values, and event status: for example, treating a promised payment as already
paid. Withholding flagged answers was not mandatory in the reported runs.

\subsection{Recorded configuration and reproducibility}
\label{sec:submitted}

The submitted 1,730-row adaptive v6 file began with
\texttt{gpt-4o-mini} \mbox{\citep{openaiGpt4oMini}} for uncertain block selection and calculation,
and \texttt{claude-sonnet-5} \mbox{\citep{anthropicSonnet5}} for identifier crops.
Tesseract processed all pages at image scale 2 and page-segmentation mode 11.
If the solver failed, it retried with
the default and then a minimal instruction profile. The structured fact list
described above was not required in this test run.

Two passes then revised answers without changing evidence IDs. First,
\texttt{gpt-4o} \mbox{\citep{openaiGpt4o}} retried 242 rows with a failed-run flag, zero answer, or at least
seven digits after the decimal point; 162 answers changed. Second, for 139 remaining zero or
long-decimal answers, Sonnet transcribed the already-selected passage crop and
\texttt{gpt-4o} solved it; 96 answers changed. Failed retries kept the previous
answer. These are change counts, not measured corrections. The triggers came
from public-training answer patterns and test outputs; a zero or decimal answer
is not necessarily wrong. Neither pass reconsidered passage selection.

Initial run records contain the passage, ranking, model settings and reply,
calculation, and answer. Revision records reconstruct the final predictions but
do not retain every crop transcription. Reconstruction is not a fresh model run.
Prompts and calculator definitions accompany the source code.

\section{Official test result}

The final leaderboard result reported for adaptive v6 was \textbf{8.61\% joint
accuracy} on 1,730 tasks. Its two revision passes left 207 answers different from
the initial run, with all 1,730 evidence fields unchanged. We have no separate
official score for each pass, so these changes do not establish an accuracy
improvement.

\paragraph{Public validation observations.}

For the 92.17\% validation run, we used \texttt{claude-opus-5}
\mbox{\citep{anthropicOpus5}} with
\texttt{xhigh} effort, a 16,000-token output limit, and one allowed attempt per task. Its prompt
required quoted facts, explicit assumptions, and a calculation. Rules selected
additional reminders about units, repetitions, prices, or event order from the
query and passage.

Table~\ref{tab:validation} records the relevant public-validation observations.
The first three rows used the released validation PDFs. The last row applies a
reading rule to previously saved model outputs, without rerunning the models or
OCR. We report the recorded scores as development observations. They are not
controlled comparisons: settings changed across runs, and the organizers
corrected validation labels during the competition.

\begin{table}[!htb]
\centering
\small
\setlength{\tabcolsep}{3pt}
\begin{tabular}{@{}lcc@{}}
\toprule
Run & Answer & Evidence F1 \\
\midrule
Two model samples & 90.32 & 1.00 \\
Combined provider outputs & 90.78 & 1.00 \\
Prompt with problem-type checks & 92.17 & 1.00 \\
Reading rule on saved outputs & 98.62 & 1.00 \\
\bottomrule
\end{tabular}
\caption{Recorded public-validation feedback (percent except evidence F1). The
first three rows came from PDF-based runs; the last recomputed saved outputs.}
\label{tab:validation}
\end{table}

The reading-rule experiment addressed a repeated error: a stated value applied
to each remaining item was interpreted as a total for those items. Recomputing
saved outputs with a rule for this wording reached 98.62\%. A separate comparison
rejected a rule that forced comparative answers to be non-negative. These were
useful error studies, but no new PDF-based run established that the revised
rules would reproduce the recorded score.

\section{Descriptive failure analysis}

We compared rendered pages with extracted blocks and saved selections after
observing failures. This manual, post-hoc analysis describes cases, not the
causal contribution of each failure mode to the official score.

\subsection{Observed document difficulties}

We inspected the first 120 test PDFs using rendered pages and their OCR
traces. Unlike the clean quantitative blocks in the public documents, inspected
test PDFs included several story-like blocks, administrative notes, repeated
topic words, stylized badges, and multiple pages.

Inspection highlighted three difficulties. Some queries supplied only broad
topical cues, such as a person, object, or activity, that matched several
passages. Some administrative notes echoed the query without containing the
requested calculation. OCR traces also contained damaged story text and badge
boundaries, while some unrelated notes remained readable. We did not count the
frequency of these patterns across the full test split.

In \texttt{task\_010565}, the query referred to a virus spreading. The rendered
page contained a virus-infection story and a printed badge. In
the scale-2 OCR trace, that story was merged into a block headed ``Purpose and
reading note,'' and its own identifier was absent. The selector's three-candidate
shortlist did not include this story. Downstream reasoning then solved an unrelated potted-flower passage. The
calculation stage could not correct this because it never received the relevant
evidence.

\subsection{Identifier reading and passage selection}

The 20-crop pilot assessed identifier transcription, not whether the selected
paragraph answered the query. Correctly transcribing the wrong paragraph's
identifier still gives incorrect evidence.

\section{Exploratory page-image study}
\label{sec:visual}

We tested a model that reads page images without the query and lists possible
word problems, their page numbers, opening words, topics, and printed identifiers.
Local code matches query words and OCR-derived topic cues to this list.
The catalog model was
\texttt{claude-haiku-4-5-}\allowbreak\texttt{20251001}
\mbox{\citep{anthropicHaiku45}}; \texttt{claude-sonnet-5} reread cropped
identifiers. Crops still required matching catalog entries to OCR block positions.
The experiment produced evidence selections, not numerical answers, and was not
a matched comparison with the submitted system.

Before running it, we sampled 100 test documents, excluding 181 used during
design or earlier inspection. The fixed sample used seed
\texttt{docsem-visual-gate-n100-v1}. The study notes describe grading by people
and assistants reading rendered pages and crops, with the system's selection
hidden during the initial passage reading. These are local judgments, not
organizer labels. The notes do not report agreement between independent graders.

\begin{table}[t]
\centering
\small
\setlength{\tabcolsep}{3pt}
\begin{tabular}{@{}p{0.65\columnwidth}r@{}}
\toprule
Measure & Result \\
\midrule
Stories selected & 55 / 100 \\
Correct story among selections & about 49 / 55 \\
Evidence emitted & 22 / 100 \\
Exact badge among emissions & about 20--21 / 22 \\
\bottomrule
\end{tabular}
\caption{Exploratory page-image study on 100 documents. Output counts are exact;
correctness estimates are approximate local judgments, not organizer labels.}
\label{tab:visualgate}
\end{table}

The recorded local estimates exceeded the declared thresholds: 80\% of selected
passages judged correct and 70\% of emitted identifiers transcribed correctly.
However, neither threshold set a minimum number of outputs. Of the 100 documents,
45 had no selected passage, 18 had a selection but no usable crop location, and
15 had conflicting identifier readings. Only 22 returned an evidence identifier.
We call this fraction coverage. Even perfect answers and identifiers for those
22 documents would give at most 22\% joint accuracy on this sample if the other
78 were left unanswered. The experiment did not measure numerical answers, and
we did not run it over the full test split.

More permissive matching on saved catalogs increased selections from 55 to 56,
but admitted administrative notes in inspected cases. This used no new model
calls and does not establish a limit for other selectors.

% Start the lessons in the next column with their framing and first item.
\newpage
\section{Lessons Learned}

These observations motivate recommendations, not demonstrated remedies.

\noindent\textbf{Check passage selection first.} The virus case motivates
inspecting the selected evidence before changing calculation instructions.

\noindent\textbf{Test varied layouts.} Compare readers or selectors on the same
documents, changing one component at a time with the solver fixed. Report
selection, identifier and answer accuracy, and coverage separately. Such
controlled comparisons are needed to establish whether a change improves
performance.

\noindent\textbf{Separate identifier and passage checks.} A correctly read
identifier can still belong to the wrong passage.

\noindent\textbf{Report coverage.} The page-image study returned evidence for
22 of 100 documents; correctness among outputs alone missed this limitation.

\section{Conclusion}

\system{} obtained 8.61\% official test joint accuracy. Our descriptive analyses
document a case of missing relevant evidence and a page-image prototype that
returned evidence for only 22 of 100 documents. They motivate testing extraction
and selection changes, but do not establish a single cause of the test result
or demonstrate that the proposed changes improve performance.

\section*{Limitations}

This report concerns one system and task. Models, prompts, readers, and metrics
differed between public validation and test experiments. The score difference
therefore cannot be attributed solely to document layout. The first 120 test
documents were a convenience sample. Correctness judgments in the 100-document
study are approximate, and raw annotation records would be needed to resolve the
remaining uncertain cases. The rule applied to saved validation outputs was
developed after inspecting errors. Its score is not a prospective test of a new
PDF-based system. Hosted model calls can also vary between runs.

\section*{Ethical considerations}

Predictions use released PDFs without private answers or source solution lookup.
A plausible calculation from an unrelated passage remains a practical risk.
Saved passages and expressions support inspection, not a guarantee of correctness.

\clearpage
\bibliography{references}
\end{document}